%% file: rtnav/root.tex
\documentclass[letterpaper, 10 pt, conference]{ieeeconf}  

\usepackage{graphicx}
\usepackage{tabularx}
\usepackage{makecell}
\usepackage{array}
\usepackage{booktabs}   
\usepackage{multirow}   
\usepackage{xcolor}
\usepackage{amsmath}
\usepackage{pifont}
\usepackage{wrapfig}
\usepackage{pifont}
\usepackage{amssymb}
\usepackage{mathtools}
\usepackage{url}
\usepackage{threeparttable}
\usepackage{adjustbox}  
\usepackage{graphicx}   
\usepackage[most]{tcolorbox}
\usepackage{xcolor}
\usepackage[table]{xcolor}
\usepackage{array}
\usepackage{graphicx}
\usepackage{capt-of}
\usepackage{fontawesome5}

\usepackage[
    colorlinks=true,
    linkcolor=black,
    citecolor=black,
    urlcolor=black,
]{hyperref}

\definecolor{promptbg}{RGB}{250,250,250}
\newtcolorbox{promptbox}[1]{
  enhanced,
  breakable,
  colback=gray!3,
  colframe=black,
  boxrule=0.4pt,
  arc=1pt,
  left=5pt,
  right=5pt,
  top=4pt,
  bottom=4pt,
  title=\textbf{#1},
  fonttitle=\footnotesize,
  coltitle=black,
  colbacktitle=gray!10
}
\definecolor{cornflowerblue}{RGB}{100,149,237}
\definecolor{coral}{RGB}{220,100,80}
\definecolor{linkblue}{RGB}{1,33,105}

\IEEEoverridecommandlockouts                              

\title{\LARGE \bf
RTNav: Towards Real-Time Zero-Shot Object Navigation
}

\author{
Easop Lee$^{1*}$,
Lingyu Zhang$^{1*}$,
and Boyuan Chen$^{1}$
}

\begin{document}

\twocolumn[{
\renewcommand\twocolumn[1][]{#1}
\maketitle

\vspace{-4mm}
\begin{center}
\small
\href{https://generalroboticslab.com/RTNav}
{\textcolor{linkblue}{\faGlobe\hspace{0.25em} Project Page}}
\qquad
\href{https://github.com/generalroboticslab/RTNav}
{\textcolor{linkblue}{\faGithub\hspace{0.25em} Code}}
\qquad
\href{https://github.com/generalroboticslab/RTNav-RealWorld}
{\textcolor{linkblue}{\faRobot\hspace{0.25em} Real-World Code}}
\end{center}
\vspace{1mm}

\begin{center}
\vspace{2mm}
    \centering 
    \vspace{0.2em}
    \includegraphics[width=\textwidth]{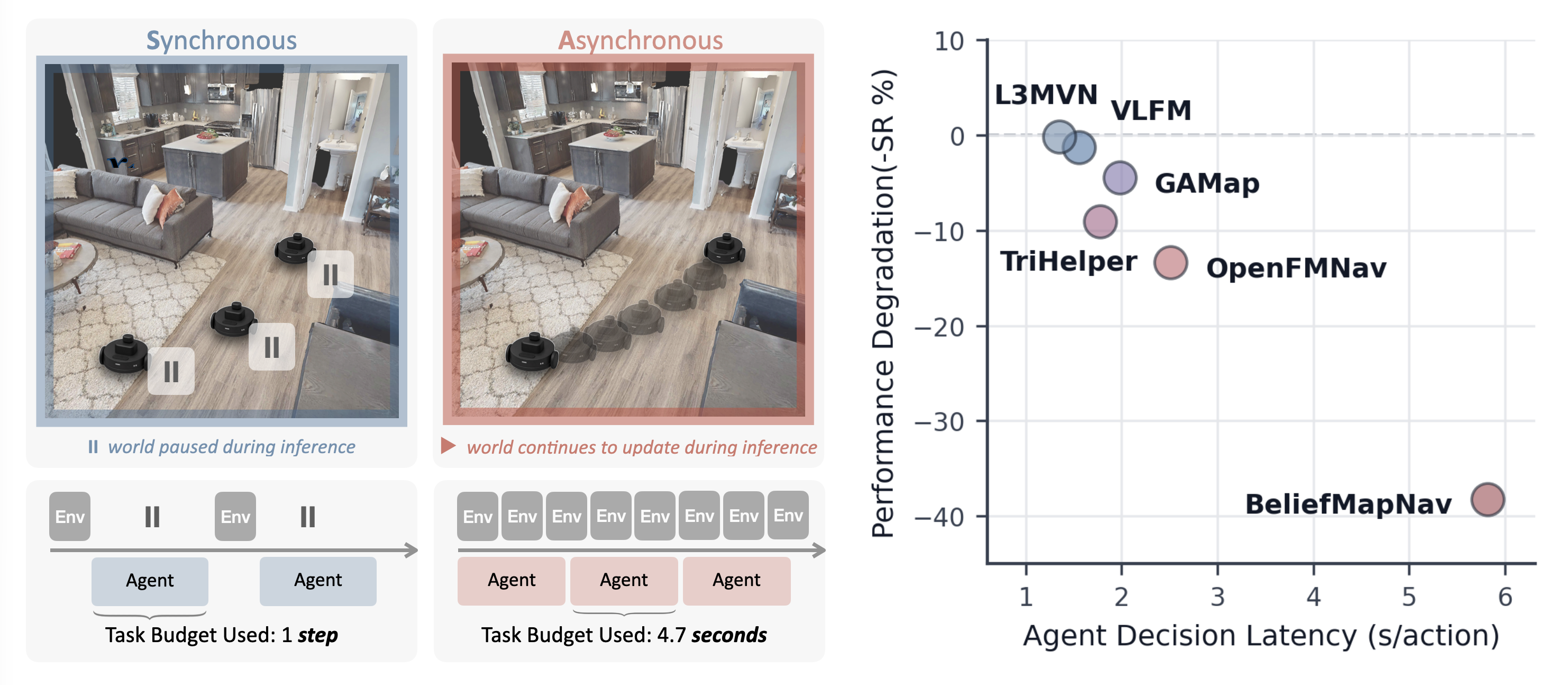}
    \captionof{figure}{\textbf{Left:} State-of-the-art navigation agents are developed in \textbf{\textit{\textcolor{cornflowerblue}{synchronous}}} environments, where the simulator is paused when the agent computes the action. This effectively allows unbounded inference times. In reality, the world updates are \textbf{\textit{\textcolor{coral}{asynchronous}}} to agent actions; it continues to evolve while the agent is running inference, and inference time costs wall-clock task budget. This fundamentally changes the task. \textbf{Right:} The performance of zero-shot navigation agents designed in synchronous environments consistently degrades under asynchronous evaluation.}
\end{center}
\vspace{5mm}
}]

\begingroup
\renewcommand{\thefootnote}{}
\footnotetext{
$^{*}$Equal contribution. $^{1}$All authors are with Duke University. This work is supported by DARPA TIAMAT program under award HR00112490419, ARO under award W911NF2410405, and ARL STRONG program under awards W911NF2320182, W911NF2220113, and W911NF242021.
}
\endgroup

\thispagestyle{empty}
\pagestyle{empty}


\begin{abstract}
Navigation in unknown environments to find unforeseen objects has become increasingly feasible with capable vision and language foundation models. However, these models also introduce non-negligible inference latency, which becomes an important concern when agents must operate continuously in the real world. Most state-of-the-art methods are still developed in synchronous simulators, where the environment waits for the agent  to act and inference time is effectively free. As a result, agents are often designed around the sequential execution of perception, reasoning, and action, with little regard for time constraints. Under real-time execution, where wall-clock time counts towards the task budget, the inefficiencies of these architectures become clear. We show that recent zero-shot object navigation methods suffer consistent performance degradation under such realistic timing conditions. Motivated by this observation, we propose RTNav, a simple but effective architecture that treats inference latency, asynchronous environment stepping, and bounded compute as explicit design considerations. Evaluated on real-time variants of HM3D-v1, HM3D-v2, and HM3D-OVON, RTNav improves the success rate by up to 11\% and the Success weighted by Completion Time by up to 5.1 points over prior work. 
\end{abstract}


\input{rtnav/sections/1_intro}
\input{rtnav/sections/2_related_work}
\input{rtnav/sections/3_method_env}

\input{rtnav/sections/4_method_agent}

\input{rtnav/sections/5_experiments}
\input{rtnav/sections/6_conclusion}


\bibliographystyle{IEEEtran}
\bibliography{rtnav/references}

\clearpage




\input{rtnav/sections/7_appendix}

\end{document}

%% file: rtnav/sections/1_intro.tex
\section{Introduction}
\label{sec:introduction}

Foundation models have become a core component of modern navigation systems ~\cite{survey}.
Large vision and language models are used to interpret goals, recognize objects, build semantic maps, and select actions without task-specific training~\cite{l3mvn, openfmnav, vlfm, trihelper, gamap, beliefmapnav}, enabling generalization to novel objects~\cite{ovon}. 
But they also introduce non-trivial inference latency. A single model query may take hundreds of milliseconds to several seconds~\cite{fastvlm}. In simulation, this delay often has little consequence. On a real robot, every second spent computing is a second in which the robot makes no progress. Slow decisions therefore directly affect responsiveness, task completion time, and ultimately whether an agent is practical to deploy.

The lack of emphasis on inference latency can be traced to the discrete, step-based simulators used for navigation evaluation. Widely-used navigation benchmarks~\cite{habitatsim, hm3d, ai2thor, procthor️} expose agents through a \textbf{\emph{synchronous}} step-based interface, in which the environment returns the next observation only \emph{after} the agent submits an action. This effectively allows agents to spend unlimited inference time per action, and standard metrics do not reflect it. Success rate is conditioned on a budget of environment steps rather than wall-lock time, and the popular Success weighted by Path Length (SPL) metric measures only the length of the trajectory. As a result, state-of-the-art methods can spend over one second per step on average waiting for a single forward pass of model outputs \cite{cognav}. On real robots, that hidden cost becomes visible as ``pause-and-go'' behavior \cite{wang2026livevln}.

Perception-action latency is a long-standing concern in robotics~\cite{nilsson1998real, tipsuwan2003control, ramstedt2019real, streaming_perception}, but it has received limited attention in modern foundation model-driven navigation systems. Synchronous benchmarks have enabled rapid prototyping and reproducible comparison, but their core assumption that inference time is negligible no longer holds at the scale of modern foundation models. 
Some methods mitigate inference delay by predicting and queuing multiple actions at once, allowing the robot to continue moving until the next prediction is available~\cite{uninav, navfom}. 
However, this still ties all modules to the same decision cycle, limiting each module from running at its full rate.
We argue that this is fundamentally an architectural problem. A synchronous architecture runs every module through a single sequential loop. The slowest module decides how reactive the whole system is. Yet, modules naturally run at different frequencies due to their input and output dimensions, sizes, memory requirements, and the number of queries. Continuous-time execution and asynchronous modularity should therefore be first-class design principles for navigation systems intended for real-world use.

To make the consequences of this design gap measurable, we introduce a real-time stepping interface that can be applied to existing object navigation benchmarks through a ROS-based layer. The simulator runs at a constant rate independently from agent action computation, while preserving the original scenes, goals, and success conditions of standard benchmarks. Existing agents can be evaluated under a real-time setting with limited modification, allowing us to study the effects of inference latency.
With this interface, we first evaluated the out-of-the-box performance of representative navigation systems on real-time variants of standard object navigation benchmarks, including HM3D-v1, HM3D-v2 \cite{hm3d} and HM3D-OVON~\cite{ovon}. To ensure baselines behave as intended, we first reproduced each method in synchronous mode--to our knowledge the largest reproducibility study in zero-shot object navigation to date. Under real-time execution, performance degrades consistently across all methods. We also find that this degradation is hardware-dependent and relates to action computation speed.

Motivated by these findings, we propose RTNav, a modular, asynchronous architecture for real-time zero-shot object navigation under bounded compute and latency constraints. Its perception, mapping, planning, and navigation modules run independently at their natural frequencies, while expensive vision-language reasoning is invoked only when needed. Across the real-time benchmarks, RTNav improves success rate by up to 11\% and Success weighted by Completion Time by up to 5.1 points over prior work.

%% file: rtnav/sections/2_related_work.tex
\section{Related Work}
\label{sec: related}

\subsection{Zero-Shot Object Navigation}

Zero-shot object navigation requires an agent to locate a specified object in previously unseen environments without additional training or adaptation at test time. 
Foundation models have enabled zero-shot generalization through open-vocabulary detection and vision-language reasoning to detect objects and guide exploration.
Earlier modular methods relied on lightweight vision-language models such as CLIP ~\cite{clip} and BLIP~\cite{blip} for semantic affinity between the goal category and visual observations to rank candidate directions ~\cite{vlfm, gamap, cows},
while more recent methods incorporate more capable, heavier LLMs and VLMs to explore~\cite{l3mvn, openfmnav, trihelper, beliefmapnav, cognav, instructnav, sgnav, voronav, wmnav, ascent}.
In parallel to the modular counterparts, end-to-end methods achieve zero-shot deployment through extensive pre-training of the VLM backbone on diverse navigation data~\cite{uninav, navfom, janusvln}. 

This shift to relying on larger, heavier foundation models or their end-to-end training variants has improved navigation success rates and path efficiency, but inference latency has grown by orders of magnitude, making wall-clock execution time a critical evaluation metric and motivating navigation systems that balance task performance with computational efficiency.

\subsection{Improving Navigation Evaluation}

The Habitat simulation ~\cite{habitatchallenge2023} is a standard platform for evaluating zero-shot object navigation, supporting benchmark datasets such as HM3D-v1, HM3D-v2, and HM3D-OVON.  
Under the standard protocol, the simulation advances \textit{synchronously} in discrete steps, waiting for the agent to return a discrete action before proceeding to the next step.
This convention reflects the historical emphasis on evaluating embodied reinforcement learning agents in an abstracted setting, rather than complete systems operating under real world deployment.
Consequently, an agent that spends 20 ms computing its next action and one that spends 2 seconds computing the identical action are evaluated identically, with wall-clock time being ignored.

Prior work has improved simulation along orthogonal directions. 
Within Habitat, VLN-CE~\cite{vlnce} replaces discrete actions with continuous control and points out the degradation that occurs when discrete navigation policies are executed in continuous environments. 
Recent work moves to Isaac simulations, introducing physically and visually more realistic environments, physics-based robot dynamics, and low-level robot control~\cite{vlnpe, 2024indoorsim,2023rethinking,truonglearning,2022bridging}.
These advances improve control fidelity and sim-to-real transfer. 

However, they do not explicitly account for wall-clock execution, leaving inference latency unaccounted for regardless of the underlying control or simulation fidelity. In this work, we preserve the standard Habitat benchmark, but expose this cost by decoupling environment progression from agent computation. 
Under this setting, we introduce RTNav, an asynchronous architecture that coordinates the full navigation stack at the system level to achieve wall-clock efficiency while preserving navigation performance.

%% file: rtnav/sections/3_method_env.tex
\section{Real-Time ObjectNav Evaluation} 
\label{sec: method_env}

\subsection{Standard ObjectNav Benchmark} 
 We first introduce the standard zero-shot ObjectNav benchmark implemented in Habitat. 
 At each decision step $k$, the agent receives an RGB-D image, GPS position, compass heading, and target object category. Based on these observations, the navigation agent predicts one of the six discrete actions: \{\textsc{MoveForward}, \textsc{TurnLeft}, \textsc{TurnRight}, \textsc{LookUp}, \textsc{LookDown}, or \textsc{Stop\}}. An episode terminates when the agent predicts \textsc{Stop} or reaches the maximum budget of 500 decision steps and is considered successful if \textsc{Stop} is issued within the \texttt{SUCCESS\_DISTANCE} of any target object viewpoint.

 Importantly, the standard benchmark follows a synchronous, step-wise execution model as follows. The agent first receives an observation and computes an action. Only after the action is passed to \texttt{env.step(action)} does the simulator execute it, advance the environment by one step, and return the next observation, after which the cycle repeats.

\subsection{Real-Time ObjectNav} 

Next, we introduce real-time ObjectNav, an evaluation setting that scores agents by task progress in wall-clock time while preserving standard benchmark tasks and datasets. Because the clock does not stop while the agent thinks, inference latency becomes part of the task: a slow agent covers less ground due to lower decision throughput. Standard evaluation hides this cost, yet this is what determines whether a method is practical on a physical robot. This evaluation benchmark thus encourages future methods to better tackle the demands of practical deployment.

The simulator is integrated using a fixed timestep of $\Delta t_{\mathrm{sim}} = 1/f = 1/30\,\mathrm{s}$, while consecutive simulation steps are paced at $1/30\,\mathrm{s}$ wall-clock intervals.
The total simulation time elapsed over an episode equals the wall-clock time.
This makes completion time a meaningful measure of performance reflected by our evaluation metrics in Section~\ref{subsec:exp_eval_metrics}.

The agent runs \textbf{\textit{asynchronously}}. Suppose inference begins at wall-clock time $t_i$ on observation $o(t_i)$ and takes $\delta_i$ seconds. The resulting command $a_i = \pi(o(t_i))$ is applied at the first simulation step after $t_i + \delta_i$, and the previous command remains active in the interim. Latency therefore degrades the system in two ways. First, the simulator advances $f\delta_i$ steps under a stale command while inference is in progress. Second, when $a_i$ is finally applied, the observation it was computed from is $\delta_i$ seconds old, so the command may no longer suit the current state. 
For instance, a decision that took one second is applied 30 simulation steps after retrieving the initial observation.

\subsection{Implementation} 

The execution model above requires the environment and the agent to run on their own paces. The simulator steps at a fixed wall-clock rate, the agent computes at its highest inference speed, and the two must interact only by exchanging observations and actions. We realize this by providing a ROS2~\cite{ros2} layer on top of Habitat. The \emph{environment node} hosts Habitat and manages episodes, publishing observations and the goal. The \emph{agent node} reads the observation and publishes velocity commands $(v, \omega) \in \mathbb{R}^2$, where $v$ is linear speed and $\omega$ is angular rate. As separate processes coupled only through observation and action topics, the simulator never waits on agent computation. Task definitions, scenes, and episodes are unchanged.

Our benchmark is designed to isolate the effect of latency. When a policy performs worse under real-time execution, that gap should be attributable to time alone, not to differences in interface or task. We therefore support two execution modes over the identical ROS2 interface. The synchronous (\texttt{sync}) mode faithfully reproduces the standard Habitat benchmark by advancing the environment only when a new action command (with updated timestamp) is received, while the asynchronous (\texttt{real}) mode advances in wall-clock time. Existing navigation methods can therefore be evaluated under either setting without modification, and comparing the same method under both modes directly quantifies the impact of latency. 
Lastly, to evaluate policies originally designed to output discrete actions under \textit{real} mode, each discrete action is mapped to an equivalent velocity (e.g., 0.25m to 0.25m/s), and executed until the corresponding translation or rotation is completed (equivalent to 1 s of wall-clock time). Details will be included in our open-source implementation.

%% file: rtnav/sections/4_method_agent.tex
\section{\texttt{RTNav}: Real-Time ObjectNav Agent}
\label{sec:method_agent}

\begin{figure*}[t]
    \vspace{0.1in}
    \centering
    \includegraphics[width=0.95\textwidth]{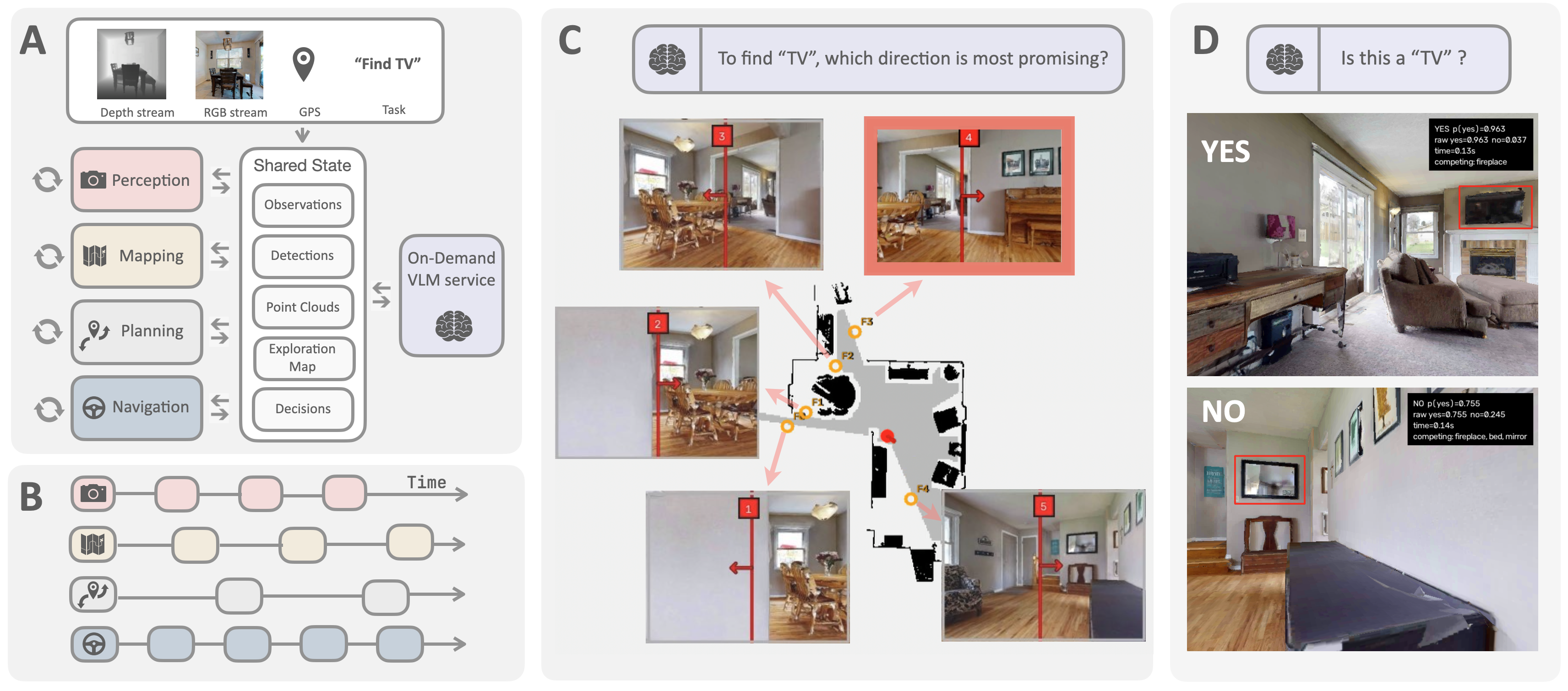}
    \caption {\textbf{RTNav Overview.} (A) The agent follows a modular, asynchronous architecture. The four main modules, perception, mapping, planning and navigation read and write into a shared state bus for communication. A VLM model operates as need-based server and aids decision-making and visual reasoning when required. (B) The modules run at their own native frequency without blocking each other. (C) During frontier selection, we simply query the VLM for most promising direction given the target object. (D) The VLM is queried when candidates are proposed by the constantly running object detector.}
    \label{fig: agent}
    \vspace{-0.9em}
\end{figure*}

In a real-time navigation system, different modules should naturally operate at different frequencies. For example, low level velocity control may run at tens of hertz, while computationally intensive subgoal planning only needs to be updated every few seconds. 

\textbf{RTNav is a navigation architecture designed for real-time deployment on a single GPU, capable of navigating to goals specified by free-form language}. As illustrated in Fig.~\ref{fig: agent}, perception, mapping, planning, and navigation modules run in independent threads and communicate through shared state. Each module reads the latest available state and publishes its output without waiting for a complete perception-reasoning-action cycle. Expensive vision-language reasoning is provided by an on-demand VLM server, allowing fast perception and control loops to continue while slower semantic reasoning is invoked only when needed.

\subsection{Modules}
\noindent \textbf{Perception.}
As the robot moves through the environment, high object detection frequency is essential to capture potential target objects. We select Owlv2~\cite{owlv2} as the open-vocabulary object detector for its state-of-the-art balance of speed and accuracy. The detector thread runs at around 15Hz on an RTX A6000 GPU. The predictions can be noisy when objects are partially occluded, viewed from difficult angles, or visually similar to other objects. We therefore use Owlv2 for high-frequency candidate proposal, and use Qwen3.5-9B~\cite{qwen35blog} for semantic verification (Fig.~\ref{fig: agent}D). Following VLFM~\cite{vlfm}, confirmed bounding boxes are further segmented by MobileSAM~\cite{mobilesam} for cleaner point cloud projection.

\noindent \textbf{Mapping.}
The mapping module maintains a 2D metric map containing obstacles, explored regions, exploration frontiers, and confirmed target locations. Incoming RGB-D observations are back-projected into an accumulated 3D point cloud of the environment, which is then projected into the 2D map. MobileSAM target masks are similarly projected to record target locations.

\noindent \textbf{Planning.} 
When no target is confirmed, the agent follows a semantically-grounded frontier exploration strategy. Frontiers selection is typically required only once several seconds, making VLM-based selection practical. For each frontier, the planner retrieves its latest associated RGB observation and marks the frontier location using a red line indicating the center of the frontier and arrow pointing towards the unexplored direction,. The VLM is then given these candidate views together with the target description and asked to select the most promising frontier, as shown in Fig.~\ref{fig: agent}C. 

\noindent \textbf{Navigation.}
Both frontier goals and target goals are reached using a PointNav policy. For fair comparison and simplicity, we adopt the same robot API described in Sec.~\ref{sec: method_env}. 
Because it runs independently, the robot can continue moving toward the latest available goal while other modules process new observations or perform semantic reasoning.

\subsection{Execution Flow}
At the beginning of each episode, RTNav receives a language description of the target object. The target is appended to a list of common household objects to form the query label set for the open-vocabulary detector. RGB and depth observations are streamed to the agent at 30~Hz and written to the shared state. The mapping module consumes the latest observations to update the point cloud, obstacle map, explored regions, and frontiers. 
We adopt the frontier extraction method from VLFM~\cite{vlfm}.
The agent begins each episode with a $360^\circ$ rotation to construct an initial map and collect an initial set of frontier observations. A new frontier is selected when the current frontier is reached, becomes invalid, or disappears as the map evolves. The VLM is therefore queried for frontier selection only when replanning is necessary. Target verification is similarly event-driven. The VLM is called only when the detector proposes a candidate. Only if the target is confirmed as true positive, it replaces the current frontier as the navigation goal.

%% file: rtnav/sections/5_experiments.tex
\section{Experiments}
\label{sec:exp}
\input{rtnav/tables/main_results}

\subsection{Evaluation Benchmarks}

We evaluated on three object navigation benchmarks: 1) HM3D-v1~\cite{hm3d} \texttt{val} split, which consists of 2000 episodes across 20 scenes with 6 object goal categories, 2) HM3D-v2~\cite{hm3d} \texttt{val} split, which consists of 1000 episodes across 20 scenes with the same 6 object goal categories, and 3) HM3D-OVON~\cite{ovon} \texttt{val\_unseen} split, which contains 3,000 episodes across 36 scenes with 49 novel object goal categories. Unlike HM3D-v1, HM3D-v2 confines itself to single-floor navigation tasks, and OVON tests the agent's open-vocabulary understanding.

\subsection{Evaluation Metrics} 
\label{subsec:exp_eval_metrics}
We report the standard Success Rate (SR), Success weighted by Path Length, and Distance to Goal (DTG) averaged over all episodes.
Path length alone cannot capture time efficiency. We therefore additionally report Success weighted by Completion Time ~\cite{yokoyama2021success}, 
\begin{equation}
    \mathrm{SCT} = \frac{1}{N}\sum_{i=1}^N S_i \, \frac{t_i^*}{\max(t_i, \, t_i^*)},
\end{equation}
where $S_i$ indicates success (0/1), $t_i$ is the wall-clock completion time of episode $i$, and $t_i^*$ is the time needed to traverse the shortest path at the robot's maximum speed. SCT thus penalizes any additional time spent on task completion, including time spent making detours and pausing for computations.

\subsection{Baselines}
We evaluated representative open-source zero-shot methods spanning a broad range of architectures, foundation models, and reasoning strategies. We limited the selection to methods with reproducible, open-sourced code that follow conventional evaluation configurations and could be evaluated within a practical time budget.
\begin{itemize}
    \item VLFM~\cite{vlfm}: constructs a BLIP-2 vision-language value map and selects frontiers with highest target relevance. 
    \item GAMap~\cite{gamap}: constructs a multi-scale CLIP value map using LLM-generated geometric-part and affordance attributes.
    \item L3MVN~\cite{l3mvn}: builds a closed-vocabulary 2D semantic map from segmentation and ranks frontiers using object co-occurrence likelihood as a commonsense prior with a
    local masked language model (RoBERTa-large).
    \item OpenFMNav~\cite{openfmnav}: extends map-based exploration to open-vocabulary targets and ranks frontiers with a proprietary LLM (\texttt{gpt-4.1-mini}) through API calls.
    \item Trihelper~\cite{trihelper}: builds on L3MVN, adding collision recovery, stalled-exploration handling, and Qwen-VL target verification.
    \item BeliefMapNav~\cite{beliefmapnav}: builds a 3D voxel belief map by combining multi-scale CLIP features with LLM-generated landmarks and uses belief-based sequential planning to select frontiers.
\end{itemize}
Because published results were obtained under the standard benchmark setting, we reran every baseline in both \texttt{sync} and \texttt{real} modes of our ROS-based simulation.
Before evaluating the baselines in \texttt{real} mode, we verified on the HM3D-v1 val-mini split that running each baseline in our \texttt{sync} mode produced the same action sequence as running its official code under the standard Habitat benchmark.
Additional details on reproduction are provided in the appendix and accompanying code.

\subsection{Implementation Details}
We evaluated all methods using commonly adopted settings for each benchmark to ensure a fair comparison.
For HM3D-v1 and HM3D-v2, we used the LocoBot with success-distance thresholds of 0.1m and 0.2m, respectively.
For HM3D-OVON, we used the Stretch robot with a 0.25m success radius.
In \texttt{real} mode, all methods were run on a single RTX A6000 GPU.
 
\subsection{Real-Time Evaluation Results}

Table.~\ref{tab:navigation_comparison_all} reports the quantitative performance of all baselines and RTNav under our unified evaluation protocol. 
Across all three benchmarks, RTNav consistently achieved the strongest overall performance, obtaining the highest success rate while maintaining competitive path efficiency and the lowest DTG. 
Under the same 500s real-time evaluation budget, RTNav surpassed the strongest baseline, which operates synchronously, by 6.4\%, 10.5\% and 9.1\% in success rate on OVON, HM3D-v1 and HM3D-v2, respectively.

RTNav achieved the most notable advantage over synchronous agents on SCT, which jointly evaluates navigation success and task completion time. 
Even against baselines with similar SPL, such as VLFM, it completed episodes in far less wall-clock time, scoring 2.45, 5.17, and 4.17 higher SCT on OVON, HM3D-v1, and HM3D-v2, respectively.
Existing methods often accumulated significant time waiting for sequential perception, reasoning, and planning modules, resulting in lower time-efficiency once computation latency is taken into account. In contrast, RTNav performed asynchronous execution, allowing perception, reasoning, mapping, and navigation to proceed concurrently while continuously controlling the robot. These results demonstrate that designing navigation systems for asynchronous execution is critical for deploying foundation model-based agents in realistic robotic settings and motivate benchmarking navigation algorithms beyond the conventional synchronous protocol.

\subsection{Real-Time Execution Analysis}

We analyzed how agents designed in synchronous environments were affected in real-time simulation and how our asynchronous design performed.

\noindent\textbf{Performance degradation under real-time evaluation.}
Comparing the sync and real columns of Table~\ref{tab:navigation_comparison_all}, we observed consistent performance degradation of baseline agents when moved from the synchronous environments they were designed in to realistic timing settings. 
Lightweight methods such as VLFM and L3MVN showed relatively small drops, whereas methods relying on computationally intensive foundation models or other modules degraded substantially more.

\begin{figure}[!t]
    \vspace{0.1in}
    \centering
    \includegraphics[width=0.45\textwidth]{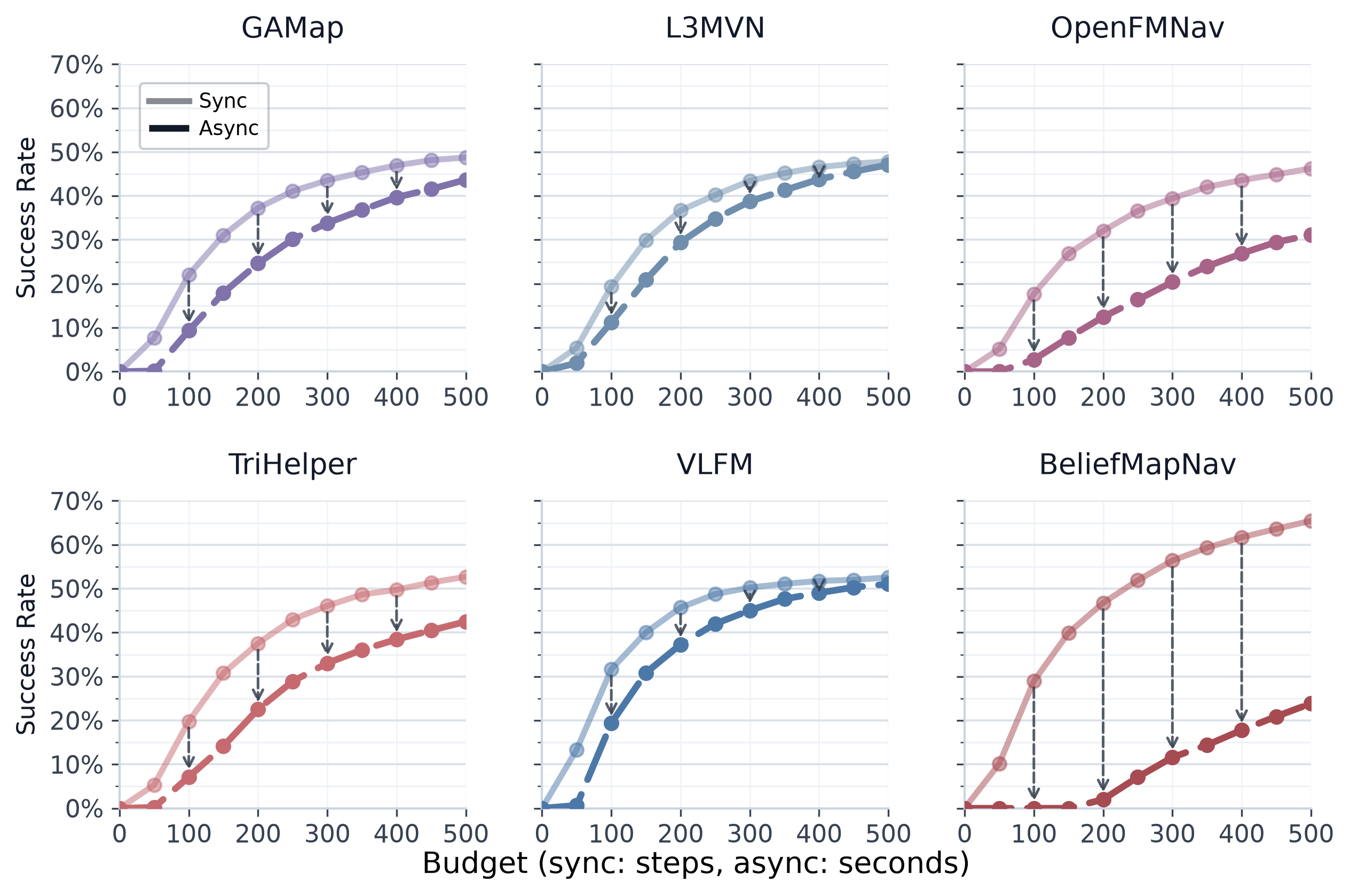}
    \caption{When agents designed in synchronous environments are evaluated under wall-clock task budget, their performance degrades consistently.}
    \label{fig: drop}
    \vspace{-0.8em}
\end{figure}

\begin{figure}[!t]
    \centering
    \vspace{0.1in}
    \includegraphics[width=0.48\textwidth]{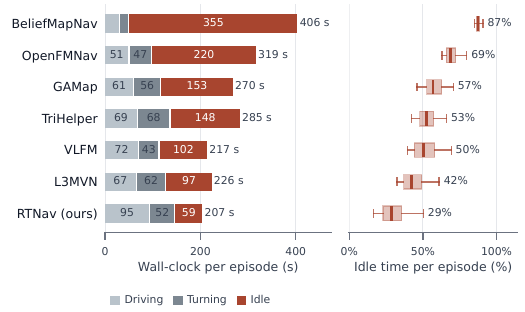}
    \caption{\textbf{RTNav showed the minimal idle time with asynchronous modules. 
    } \textbf{Left:} Total completion time comprises of the driving time, turning time, and idle time. Idle time consists of any time the agent's pose is stationary including latency, being stuck due to collision, and moving camera up/down. \textbf{Right:} RTNav showed 14\%+ less idle time compared to all other baselines.}
    \label{fig:idletime}
\end{figure}

\begin{figure}[!t]
    \includegraphics[width=0.48\textwidth]{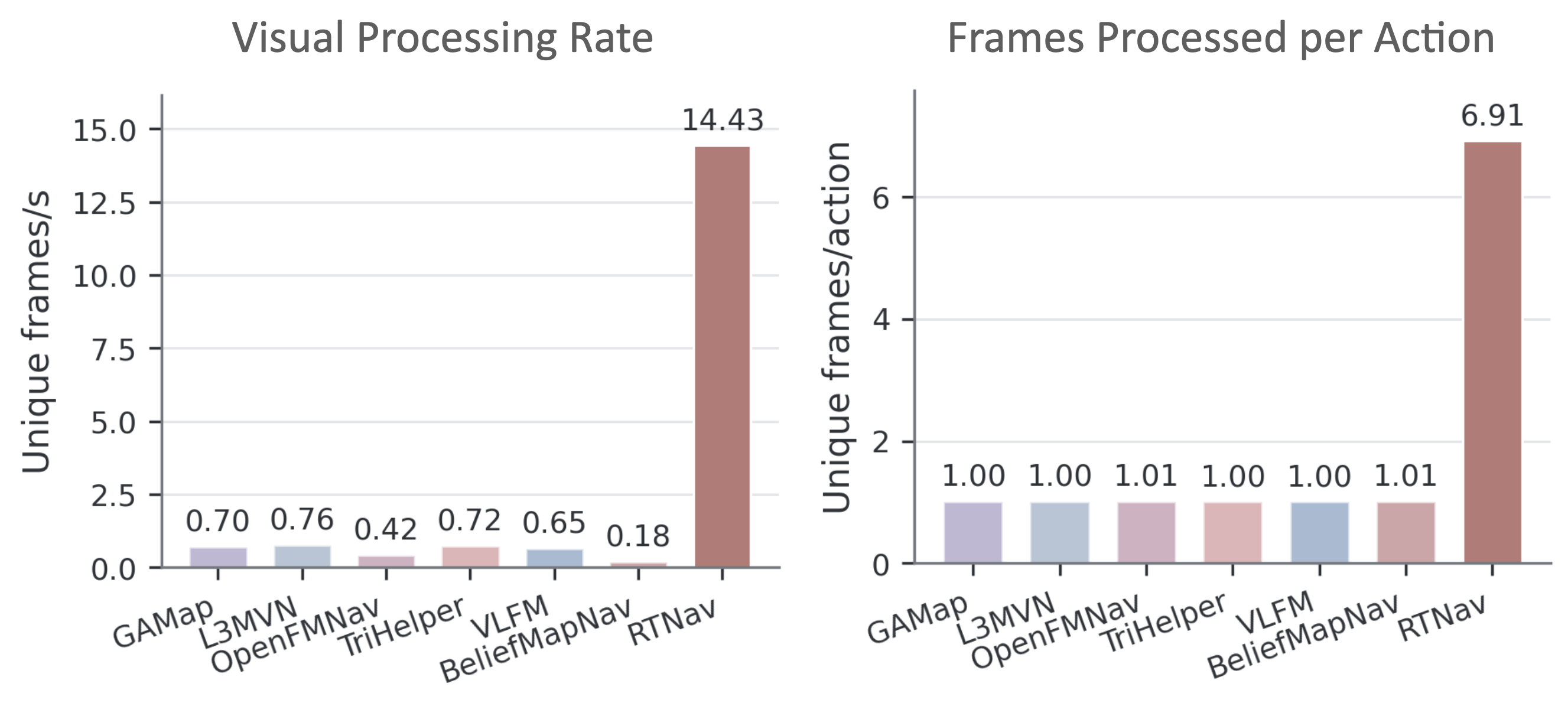 }
    \caption {\textbf{Asynchronous architecture enabled significantly higher visual processing speed.} \textbf{Left:} Without other modules blocking, RTNav reached over $20\times$ unique detection frames per second than synchronous baselines. \textbf{Right:} RTNav processed more detection frames per actions, where synchronous agents are limited to around 1 detection per action.}
    \label{fig:frames}
\end{figure}

\begin{figure}[!t]
    \includegraphics[width=0.44\textwidth]{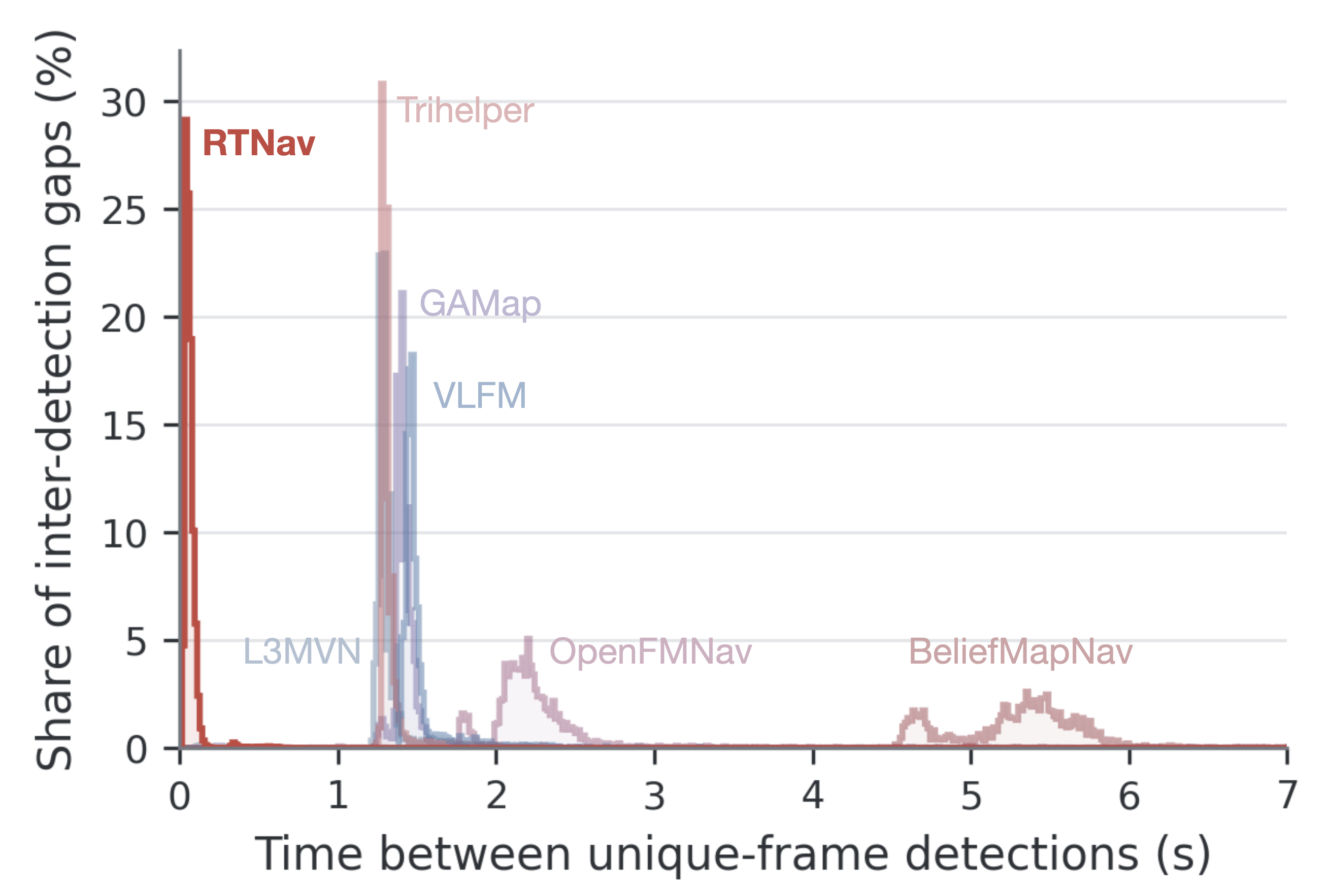}
    \caption {\textbf{Blind gaps between detections were significantly shorter in RTNav.} Asynchronous architecture missed less visual information.}
    \label{fig:histogram}
    \vspace{-0.8em}
\end{figure}

\begin{figure*}[!t]
    \vspace{0.1in}
    \centering
    \includegraphics[width=\textwidth]{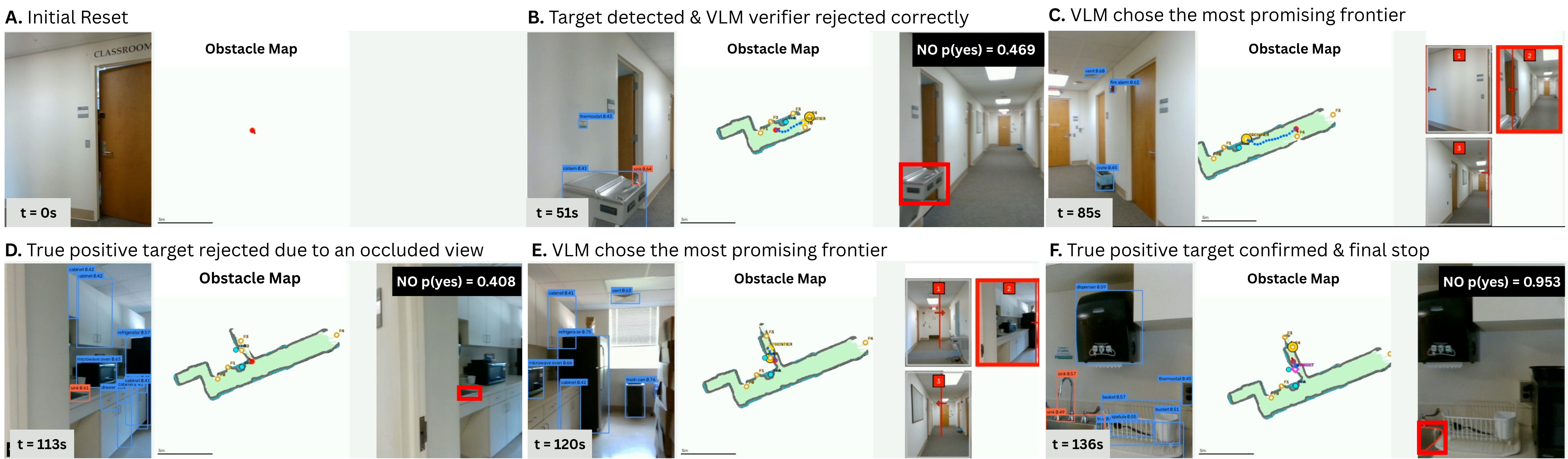}
    \caption{Representative snapshots from a successful trajectories of the RTNav agent in the real world. Target object: sink.
    }
    \label{fig:qualitative}
    \vspace{-0.8em}
\end{figure*}

Furthermore, Fig.~\ref{fig: drop} traces how this degradation accumulates over the episode budget. We plotted \texttt{sync} mode and \texttt{real} mode success rate against episode budget (500 steps for \texttt{sync}, 500 seconds for \texttt{real}) for all baselines. The gap between the curves at each budget (arrows) represents the success lost once evaluation reflects real-time execution. For fast methods, the drop was relatively minor, indicating that the time they spent on inference was not significant enough to severely impact success rate given a time budget. For slower methods, the real-time curve plateaued further below the synchronous one. These results demonstrate that the conventional step-wise evaluation protocol can substantially overestimate the practical performance of modern foundation model-based navigation systems.

\noindent\textbf{Minimizing idle time through asynchronous execution.}
Fig.~\ref{fig:idletime} compares the execution timeline of synchronous and asynchronous navigation. Under synchronous execution, the robot repeatedly waits for all modules to complete before issuing the next control command, resulting in significant idle time. In contrast, RTNav executes these modules concurrently, allowing the robot to continue navigating while higher-level modules process incoming observations. This substantially reduced idle time and wall clock task completion time.

\input{rtnav/tables/hardware_comparison}
\input{rtnav/tables/vlm_ablation}

\noindent\textbf{Seeing in real-time.} Without other modules blocking, the perception thread in RTNav achieved a significantly higher detection frequency, reaching over $20 \times$ unique detection frames per second than other baselines (Fig.~\ref{fig:frames}). With higher number of unique frames, more visual evidence can be collected to mitigate the noisiness of imperfect object detections. This also allowed more frames to be processed between actions, where synchronously-designed baselines are inherently limited at 1 frame per action. Consequently, RTNav misses far less visual information. As shown in Fig.~\ref{fig:histogram}, the inter-detection gaps were an order of magnitude shorter than synchronous baselines. 

\subsection{Sensitivity Analysis}

We further studied the sensitivity of RTNav to compute hardware and the choice of vision-language model. Table ~\ref{tab:hardware} compared three compute tiers: the RTX A6000 used in our main experiments, the L40S as an alternative for offboard compute, and the Jetson Thor representing edge compute. RTNav achieved nearly identical SR and SPL across all three platforms, while SCT decreased moderately on the Jetson Thor. These results suggested that RTNav was robust to substantial changes in compute hardware, although slower edge compute still affected wall-clock efficiency.

We evaluated three open-source VLMs with varying sizes: Qwen3.5-4B, Qwen3.5-9B, and Gemma4-12B. 
Table~\ref{tab:vlm} shows Qwen3.5-9B achieved the best overall performance, slightly outperforming the larger Gemma4-12B. In contrast, reducing the model to Qwen3.5-4B substantially degraded all metrics. These results suggested that RTNav did not require the largest available VLM, but sufficient semantic reasoning capacity remained important for reliable frontier selection.

\subsection{Qualitative Results}
Fig. ~\ref{fig:qualitative} shows an example episode progress for RTNav in the real world. Semantic frontier selection guides exploration, while high-frequency detection captures peripheral targets and VLM verification filters false positives. Additional simulation and real-world results are included in the supplementary video, and full implementation details can be found in the open-source code.

%% file: rtnav/tables/main_results.tex
\newcommand{\venue}[1]{\textcolor{gray}{\scriptsize\text{#1}}}

\begin{table*}[!t]
\vspace{0.15in}
\caption{Navigation performance across benchmarks. SR = Success Rate, SPL = Success weighted by Path Length, DTG = Distance to Goal, SCT = Success weighted by Completion Time. Bold indicates best performance per metric within each benchmark.}
\vspace{\baselineskip}
\centering
\scalebox{0.88}{
\small
\begin{tabular}{lllccccccccc>{\columncolor{blue!5}}c}
\toprule
Benchmark & Method & Venue &\multicolumn{3}{c}{SR $\uparrow$} & \multicolumn{3}{c}{SPL $\uparrow$} & \multicolumn{3}{c}{DTG $\downarrow$} & SCT $\uparrow$ \\
\cmidrule(lr){4-6} \cmidrule(lr){7-9} \cmidrule(lr){10-12}
& & & \texttt{sync} & \texttt{real} & $\Delta$ & \texttt{sync} & \texttt{real} & $\Delta$ & \texttt{sync} & \texttt{real} & $\Delta$ & \texttt{real} \\
\midrule
\multirow{2}{*}{OVON}
& VLFM~\cite{vlfm} & \venue{ICRA'24}
& $39.8$ & $38.2$ & \textcolor{red!80}{\scriptsize$-1.6$}
& $24.2$ & $\textbf{23.7}$ & \textcolor{red!80}{\scriptsize$-0.5$}
& $3.50$ & $3.63$ & \textcolor{red!80}{\scriptsize$+0.13$}
& $5.37$ \\
& \textbf{RTNav (Ours)} & \venue{}
& - & $\mathbf{44.6}$ & -
& - & ${21.1}$ & -
& - & \textbf{3.36} & -
& $\mathbf{7.82}$ \\
\midrule
\multirow{7}{*}{HM3D-v1}
& L3MVN~\cite{l3mvn} & \venue{IROS'23}
& $47.9$ & $47.7$ & \textcolor{red!80}{\scriptsize$-0.2$}
& $22.6$ & $22.5$ & \textcolor{red!80}{\scriptsize$-0.1$}
& $4.62$ & $4.61$ & \textcolor{green!55!black!50}{\scriptsize$-0.02$}
& $7.29$ \\
& VLFM~\cite{vlfm} & \venue{ICRA'24}
& $52.6$ & $51.3$ & \textcolor{red!80}{\scriptsize$-1.3$}
& $30.4$ & $30.0$ & \textcolor{red!80}{\scriptsize$-0.4$}
& $4.12$ & $4.21$ & \textcolor{red!80}{\scriptsize$+0.09$}
& $8.95$ \\
& OpenFMNav~\cite{openfmnav} & \venue{NAACL-F'24}
& $46.3$ & $32.9$ & \textcolor{red!80}{\scriptsize$-13.4$}
& $20.7$ & $17.3$ & \textcolor{red!80}{\scriptsize$-3.4$}
& $4.78$ & $5.45$ & \textcolor{red!80}{\scriptsize$+0.66$}
& $2.99$ \\
& Trihelper~\cite{trihelper} & \venue{IROS'24}
& $52.7$ & $43.7$ & \textcolor{red!80}{\scriptsize$-9.1$}
& $23.7$ & $20.5$ & \textcolor{red!80}{\scriptsize$-3.1$}
& $4.19$ & $4.70$ & \textcolor{red!80}{\scriptsize$+0.51$}
& $5.30$ \\
& GAMap~\cite{gamap} & \venue{NeurIPS'24}
& $48.8$ & $44.3$ & \textcolor{red!80}{\scriptsize$-4.5$}
& $23.6$ & $23.3$ & \textcolor{red!80}{\scriptsize$-0.4$}
& $4.58$ & $4.85$ & \textcolor{red!80}{\scriptsize$+0.27$}
& $5.67$ \\
& BeliefMapNav~\cite{beliefmapnav} & \venue{NeurIPS'25}
& $65.5$ & $27.3$ & \textcolor{red!80}{\scriptsize$-38.2$}
& $32.2$ & $18.7$ & \textcolor{red!80}{\scriptsize$-13.5$}
& $2.95$ & $5.38$ & \textcolor{red!80}{\scriptsize$+2.44$}
& $1.53$ \\
& \textbf{RTNav (Ours)} & \venue{}
& - & \textbf{61.8} & -
& - & \textbf{30.4} & -
& - & \textbf{3.50} & -
& \textbf{14.12} \\
\midrule
\multirow{7}{*}{HM3D-v2}
& L3MVN~\cite{l3mvn} & \venue{IROS'23}
& $58.9$ & $58.3$ & \textcolor{red!80}{\scriptsize$-0.6$}
& $24.9$ & $25.5$ & \textcolor{green!55!black!50}{\scriptsize$+0.6$}
& $2.51$ & $2.46$ & \textcolor{green!55!black!50}{\scriptsize$-0.04$}
& $8.12$ \\
& VLFM~\cite{vlfm} & \venue{ICRA'24}
& $64.0$ & $63.2$ & \textcolor{red!80}{\scriptsize$-0.8$}
& $32.7$ & \textbf{32.7} & \textcolor{black!50}{\scriptsize$+0.0$}
& $2.01$ & $2.06$ & \textcolor{red!80}{\scriptsize$+0.05$}
& $9.67$ \\
& OpenFMNav~\cite{openfmnav} & \venue{NAACL-F'24}
& $57.5$ & $45.1$ & \textcolor{red!80}{\scriptsize$-12.4$}
& $23.8$ & $21.7$ & \textcolor{red!80}{\scriptsize$-2.1$}
& $2.43$ & $3.11$ & \textcolor{red!80}{\scriptsize$+0.68$}
& $3.65$ \\
& Trihelper~\cite{trihelper} & \venue{IROS'24}
& $64.0$ & $57.2$ & \textcolor{red!80}{\scriptsize$-6.8$}
& $26.0$ & $24.3$ & \textcolor{red!80}{\scriptsize$-1.8$}
& $1.98$ & $2.27$ & \textcolor{red!80}{\scriptsize$+0.29$}
& $6.07$ \\
& GAMap~\cite{gamap} & \venue{NeurIPS'24}
& $61.3$ & $57.4$ & \textcolor{red!80}{\scriptsize$-3.9$}
& $27.1$ & $26.5$ & \textcolor{red!80}{\scriptsize$-0.6$}
& $2.32$ & $2.43$ & \textcolor{red!80}{\scriptsize$+0.12$}
& $6.11$ \\
& BeliefMapNav~\cite{beliefmapnav} & \venue{NeurIPS'25}
& $73.5$ & $36.7$ & \textcolor{red!80}{\scriptsize$-36.8$}
& $33.6$ & $22.4$ & \textcolor{red!80}{\scriptsize$-11.2$}
& $1.79$ & $3.23$ & \textcolor{red!80}{\scriptsize$+1.44$}
& $1.85$ \\
& \textbf{RTNav (Ours)} & \venue{}
& - & \textbf{72.3} & -
& - & $30.8$ & -
& - & \textbf{1.38} & -
& \textbf{13.84} \\
\bottomrule
\end{tabular}
}
\label{tab:navigation_comparison_all}
\vspace{-0.5em}
\end{table*}

%% file: rtnav/tables/hardware_comparison.tex
\begin{table}[t]
\caption{Sensitivity of RTNav to compute hardware.}
\label{tab:hardware}
\centering
\footnotesize
\begin{tabular*}{0.82\columnwidth}{@{\extracolsep{\fill}}lccc@{}}
\toprule
\textbf{Hardware} & \textbf{SR} $\uparrow$ & \textbf{SPL} $\uparrow$ & \textbf{SCT} $\uparrow$ \\
\midrule
RTX A6000   & 72.3 & 30.8 & 13.84 \\
L40S        & 72.2  & 31.5  & 14.13   \\
Jetson Thor & 72.2  & 31.4  & 12.73   \\
\bottomrule
\end{tabular*}
\end{table}

%% file: rtnav/tables/vlm_ablation.tex
\begin{table}[!t]
\caption{Effect of the VLM used for semantic frontier selection.}
\label{tab:vlm}
\centering
\footnotesize
\begin{tabular*}{0.82\columnwidth}{@{\extracolsep{\fill}}lccc@{}}
\toprule
\textbf{VLM} & \textbf{SR} $\uparrow$ & \textbf{SPL} $\uparrow$ & \textbf{SCT} $\uparrow$ \\
\midrule
Gemma4-12B & 66.2 & 29.9 & 13.45 \\
Qwen3.5-4B & 37.8 & 10.5 & 4.56 \\
Qwen3.5-9B & \textbf{72.3} & \textbf{30.8} & \textbf{13.84} \\
\bottomrule
\end{tabular*}
\vspace{-0.5em}
\end{table}

%% file: rtnav/sections/6_conclusion.tex
\section{Conclusion}

Common synchronous navigation benchmarks are not realistic reflections of real-time deployment as they ignore agent computation time. 
We show that recent foundation model-driven object navigation methods' performance consistently degrades once inference latency counts towards the task budget, with slower methods suffering the largest drops. Motivated by this gap, we introduce \texttt{RTNav}, a modular asynchronous agent in which perception, mapping, planning, and navigation operate independently at native frequencies. Across HM3D-v1, HM3D-v2, and HM3D-OVON, RTNav achieves strong real-time performance, improving SR by up to 11\% and SCT by up to 5.1 points over prior methods. These results highlight the importance of treating wall-clock efficiency and asynchronous execution as first-class considerations when designing navigation systems for real-world deployment.

There are several limitations to our study. 
Extending asynchronous, wall-clock-aware design to more tightly coupled embodied tasks, such as locomotion and manipulation, is an important direction for future work. 
We also primarily consider static environments, leaving dynamic scenes for future study.

%% file: rtnav/sections/7_appendix.tex
\section*{APPENDIX}

\subsection{Real-Time Environment Implementation}

\begin{figure}[h]
    \centering
    \includegraphics[width=\linewidth]{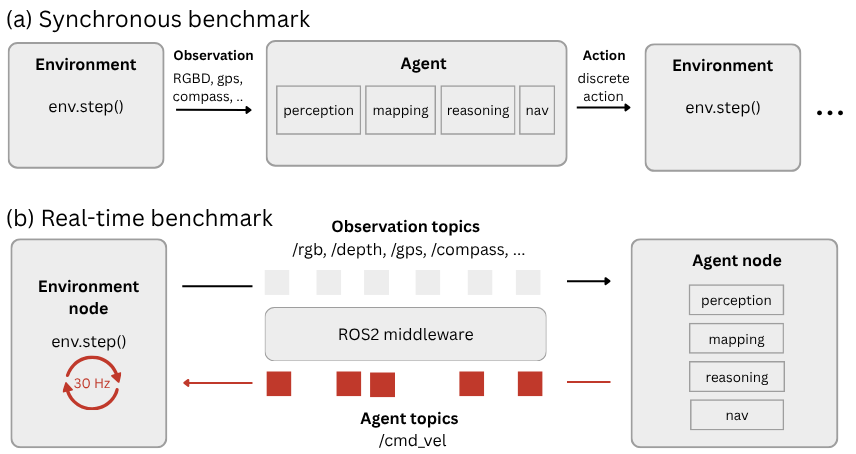}
    \caption{Synchronous benchmark waits for the agent output indefinitely for the next simulation stepping, whereas the real-time benchmark runs both environment and agent node separately in their own frequencies.}
    \label{fig:env_ros}
\end{figure}

As shown in Fig.~\ref{fig:env_ros}, our simulation benchmark was implemented through a ROS-based interface that supports both the original synchronous evaluation protocol and asynchronous real-time execution. In synchronous mode, \texttt{env.step()} was called whenever the agent produces a new action. In real-time mode, it was called continuously at 30Hz using the latest available action and an updated timestamp.

The agents in the synchronous setting output discrete actions: \{\textsc{MoveForward}, \textsc{RotateLeft}, \textsc{RotateRight}, \textsc{LookUp}, \textsc{LookDown}, or \textsc{Stop\}}. To adapt these discrete agents to our real-time simulation design, we converted each discrete action to its corresponding continuous command as shown in the following table:

\input{rtnav/tables/action_map}

With this action  mapping, when inference is instant, 500 steps in synchronous mode and 500 seconds in asynchronous mode task budgets are effectively equivalent. We then send these velocity commands to the simulator and continuously checked at every simulation step (@30 Hz) if the termination conditions were met.

\subsection{Baseline Implementation Details}
\label{sec:appendix_baselines}

We made our best effort to reproduce the original reported results in synchronous mode to ensure baseline methods reached its intended performance. Below, we document details on how we handled missing implementation details, discrepancies between the released code and papers, and outdated or deprecated dependencies. 

\noindent \textbf{L3MVN} The official repository (\url{https://github.com/ybgdgh/L3MVN/}) was tested on \texttt{torch=1.7.0}, \texttt{torchvision=0.8.0} and \texttt{python=3.7}. However, Habitat version used in our evaluation requires \texttt{python=3.9}. We used the closest compatible versions,
\texttt{torch=1.9.1} and \texttt{torchvision=0.9.0}. We followed the paper's description of using RoBERTa instead of GPT-2 in the official released code.

\noindent \textbf{OpenFMNav}
The official OpenFMNav repository (\url{https://github.com/yxKryptonite/OpenFMNav}) was tested with \texttt{Python 3.9.16} and \texttt{PyTorch 1.11.0+cu113}. Since our machine used CUDA 12.2, we used the closest compatible versions, \texttt{PyTorch 2.1.1}, \texttt{torchvision 0.16.1}, and \texttt{pytorch-cuda=12.1}.
The models \texttt{gpt-4-1106-preview} and \texttt{gpt-4-vision-preview} models used in the original paper are deprecated. We replaced them with comparable, if not stronger and faster \cite{openai_simple_evals_benchmarks}, models: \texttt{gpt-4.1-mini} and \texttt{gpt-4o}.

\noindent \textbf{VLFM}
The original VLFM paper used YOLO for in-coco-class detection and GroundingDINO as a novel category detector. Following OVON\cite{ovon}, we replace GroundingDINO with the better-performing Owlv2\cite{owlv2}. The specific variant of Owlv2 was not specified in the paper or released code. We used the same \texttt{owlv2-base-patch16-ensemble} as in RTNav.

\noindent \textbf{TriHelper}
We used the authors’ released implementation. When agent configurations conflicted between the code and the paper, we selected the paper's specified configurations. 

\noindent \textbf{GAMap} The official GAMap repository was tested with \texttt{Python 3.9.16} and \texttt{PyTorch 1.11.0+cu113}. Since our machine used CUDA 12.2, we used the closest compatible versions, \texttt{PyTorch 2.1.0}, \texttt{torchvision 0.16.0}, and \texttt{pytorch-cuda=12.1}.

\input{rtnav/tables/reprod}

\noindent \textbf{BeliefMapNav} The original implementation used Habitat 0.2.5. To ensure that all methods were evaluated under the same simulator version, we re-evaluated BeliefMapNav using Habitat 0.2.4. 

We made several minor implementation fixes required for robust large-scale evaluation, without modifying the underlying navigation method. In particular, we handled failure cases that would otherwise stall the evaluation: 1) treated malformed VLM verification responses that did not contain the expected yes/no output as positive responses, and 2) constrained the maintained map to the predefined spatial bounds when map updates extended beyond the allocated region. These changes were made solely to prevent implementation failures that interrupted evaluation and allow evaluation to proceed across all episodes.

For methods using the SemExp planner (L3MVN, OpenFMNav, TriHelper, and GAMap), we computed pose changes directly from the simulator's GPS and compass readings rather than assuming ideal motion from commanded actions. This prevented unintended map drift caused by any asynchronous execution and ROS communication delay.

We verified the correctness of our ROS-wrapped baselines by running both our eval system (sync mode) and the official codebases on the 30-episode \texttt{val\_mini} split of HM3D-v1, and checked that all actions matched exactly. For BeliefMapNav, exact action-level matching was difficult due to GPU-accelerated Open3D operations in the original implementation. In Table~\ref{tab:reprod}, we compare our reproduced results using our ROS evaluation stack in sync mode with the reported results.

\subsection{RTNav Implementation Details}

\paragraph{Common objects/synonyms} We constructed the detector vocabulary from a subset of LVIS~\cite{lvis} containing 405 common indoor object categories. We excluded six small fixture categories: \textit{doorknob}, \textit{handle}, \textit{hinge}, \textit{knob}, \textit{latch}, and \textit{wall socket}. At the beginning of each episode, the VLM first generates up to two common synonyms/aliases for the given target category. The generated synonyms and the filtered LVIS categories are then passed to the \texttt{google/embeddinggemma-300m} model to reduce the candidate set to a small subset of potentially relevant categories. We then query the VLM to determine which of these candidates are true synonyms to the target object. 

\paragraph{Perception and mapping.}
RTNav uses the \texttt{owlv2-base-patch16-ensemble} detector with a $960\times960$ inference resolution and a confidence threshold of 0.25. The detector evaluates the target aliases together with a fixed 399-category indoor/LVIS vocabulary. Seven text templates are ensembled for each category, overlapping boxes are suppressed using an NMS IoU threshold of 0.6.

\paragraph{Object memory and verification.}
Each OWLv2 box is back-projected with depth to form a 3D detection. A persistent object graph is maintained for each episode. A object node becomes confirmed after observations from three distinct viewpoints; two observations count as distinct when the camera translation differs by at least 0.1\,m or its yaw differs by approximately $3^\circ$.

\begin{figure*}[h]
    \centering
    \includegraphics[width=\linewidth]{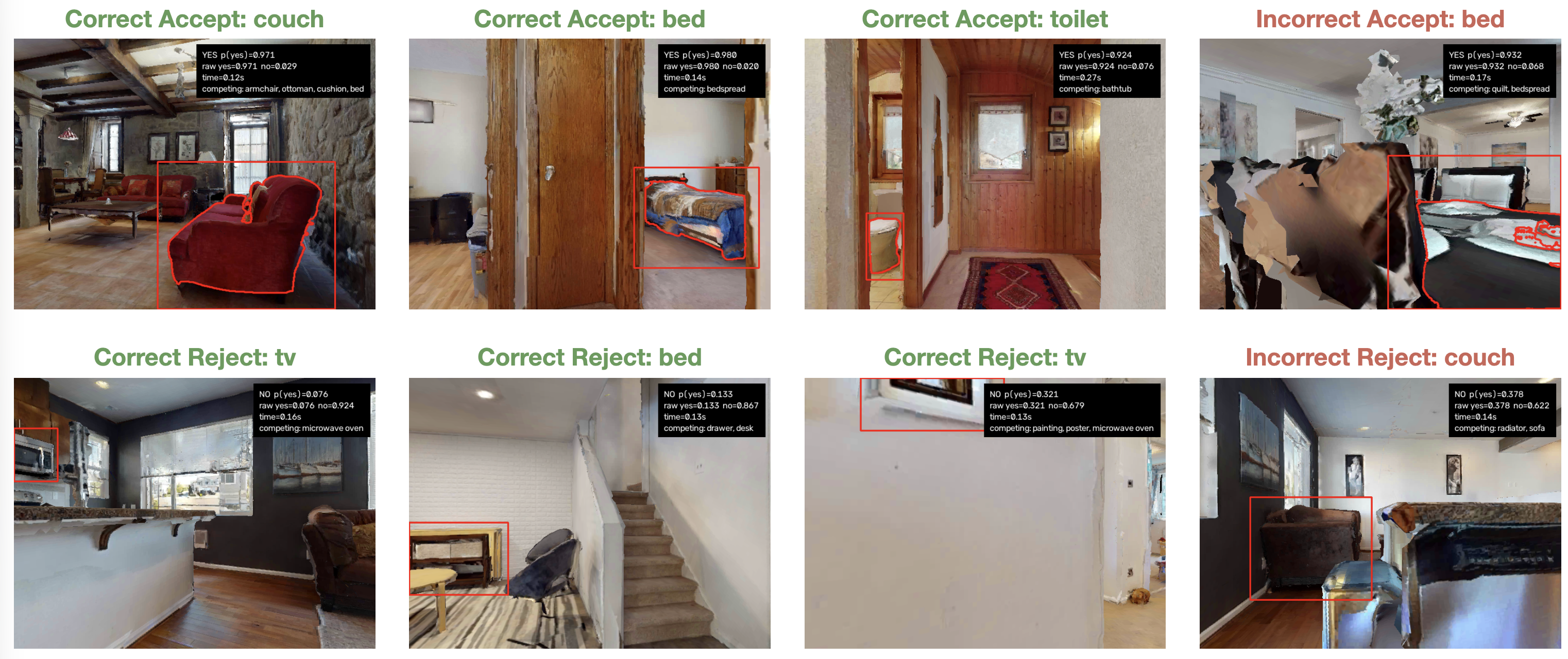}
    \caption{\textbf{Qualitative examples of VLM object verfication}.}
    \label{fig:qual_vlm}
\end{figure*}

A target candidate is shown to the VLM with an expanded red bounding box and a one-token yes/no prompt. Up to eight candidates are submitted in parallel. RTNav computes
\[
p_{\mathrm{yes}} =
\frac{p(\text{yes})}{p(\text{yes})+p(\text{no})}
\]
from the first-token log probabilities and accepts a candidate only when $p_{\mathrm{yes}}>0.90$. Competing labels from the detector or object graph are included in the prompt when available. An object node is blacklisted after three confident rejections with $p_{\mathrm{yes}}\leq0.50$. Qualitative examples of this step are shown in Fig.~\ref{fig:qual_vlm}.

After VLM acceptance, we follow VLFM and use MobileSAM to refine the object mask. The mask is projected into 3D to obtain both an object centroid and a nearest surface navigation goal. When detected objects are more than 3 meters away from the robot or the mask is truncated by the image boundary, the target is marked as candidate target. RTNav approaches such a candidate to acquire a better view. If it is verified again within 3 meters, it becomes confirmed target for the episode. If it was failed to be re-verified, it is abandoned as a target. 

\paragraph{Implementation efficiency}
RTNav executes perception, mapping, frontier detection, decision-making, and navigation asynchronously. To improve the compute efficiency, we cache the text embeddings of the fixed detector vocabulary and perform batched inference across camera views. GPU-intensive models are coordinated to avoid memory contention, while batched C++ extensions accelerate detection post-processing and depth-based 2D-to-3D projection.

\paragraph{VLM prompts} The VLM is called on demand for synonym selection, frontier selection and object verification. The prompts used for these calls are provided below.

\begin{promptbox}{Initial Synonym Generation}
\footnotesize\ttfamily
Given one indoor object, return a Python list containing zero to two common,
unambiguous alternative names for exactly the same visible physical object.
Do not return rooms, associated objects, or explanations. Object: \{target\}
\end{promptbox}

\begin{promptbox}{Candidate-synonym verification}
\footnotesize\ttfamily
A robot is asked to search for: \{target\}, the indoor object, in a navigation task.
It detected the object \{candidate\}. The detector is not perfect: it might label
the target as the candidate if they are visually similar, but it might also label
an incorrect object as the candidate if only a small part looks similar. Consider
the user's intent in indoor object navigation. Would the user be satisfied with
\{candidate\}? Respond with exactly one word: yes or no.
\end{promptbox}

\begin{promptbox}{Frontier Selection}
\footnotesize\ttfamily
You are a navigation selector. Reply with only one allowed integer and no words.

You are selecting a frontier for ObjectNav.

Target: \{target\}

Each numbered panel is an independent candidate. Red numbers,
lines, and arrows are navigation overlays, not scene objects. The arrow indicates
the candidate direction.

Choose the candidate with the highest expected progress toward the target:
- Prefer most semantically relevant direction if there is one.
- Otherwise, prefer open, traversable areas that has high potential leading to a semantically
  plausible room;
- Avoid walls, closed doors, dead ends, stairs;

Return exactly one valid direction number between 1-\{len(entries)\} and no other text.
\end{promptbox}

\begin{promptbox}{Object Verification}
\footnotesize\ttfamily
Look at the object inside the red bounding box. 
Question: is this object a \{target\}?
Note: similar-looking objects may include \{competing\_label\}.
Answer yes only if the boxed object is clearly the target object.
Return exactly one word: yes or no.
\end{promptbox}

\subsection{Task Details}

We evaluated all methods in Habitat using the official HM3D-v1, HM3D-v2, and OVON validation splits. The simulator configurations are summarized in Table~\ref{tab:task-details}. Sliding was disabled for all benchmarks. The agent has forward step size of $0.25$\,m and a turn angle of $30^\circ$. An episode was considered a success when the agent called \textsc{Stop} within the benchmark-specific success distance of a target instance.

\begin{table}[h]
\centering
\caption{Task and simulator configurations.}
\label{tab:task-details}
\small
\begin{tabular}{lccc}
\toprule
 & HM3D-v1 & HM3D-v2 & OVON \\
\midrule
Evaluation split      & val & val & val-unseen \\
Episodes              & 2,000 & 1,000 & 3,000 \\
Target categories     & 6 & 6 & 49 \\
Robot height (m)      & 0.88 & 0.88 & 1.41 \\
Robot radius (m)      & 0.18 & 0.18 & 0.17 \\
Camera height (m)     & 0.88 & 0.88 & 1.31 \\
RGB/depth resolution  & $640{\times}480$ & $640{\times}480$ & $360{\times}640$ \\
Horizontal FoV        & $79^\circ$ & $79^\circ$ & $42^\circ$ \\
Depth range (m)       & $[0.5,5.0]$ & $[0.5,5.0]$ & $[0.5,5.0]$ \\
Success distance (m)  & 0.10 & 0.20 & 0.25 \\
Task budget(steps/s)       & 500 & 500 & 500 \\
Forward step (m)      & 0.25 & 0.25 & 0.25 \\
Turn angle            & $30^\circ$ & $30^\circ$ & $30^\circ$ \\
Sliding               & disabled & disabled & disabled \\
\bottomrule
\end{tabular}
\end{table}

\begin{figure*}
    \centering
    \includegraphics[width=\linewidth]{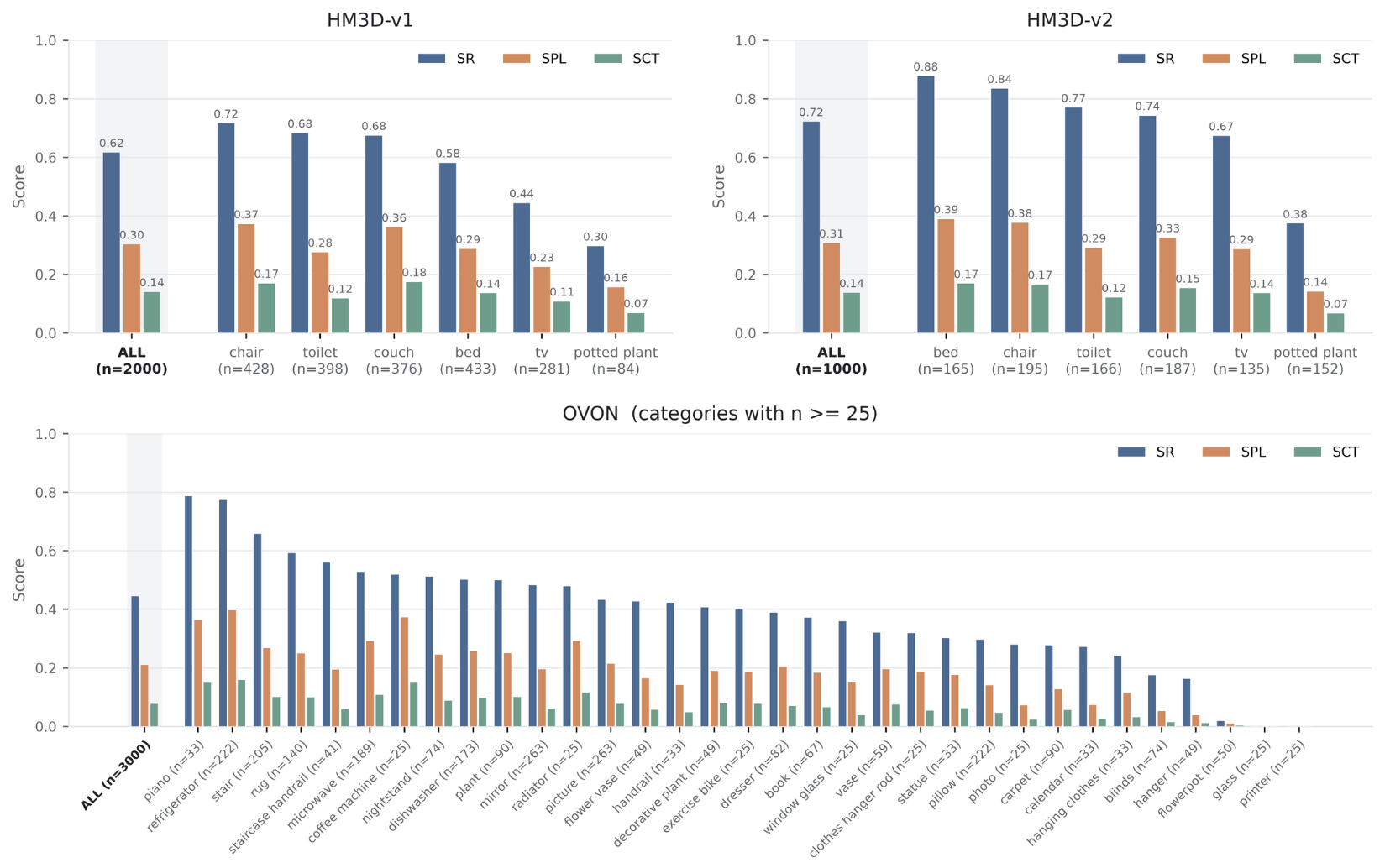}
    \caption{Category-level success rate (SR), success weighted by path length
(SPL), and success weighted by completion time (SCT) for RTNav. OVON
categories with fewer than 25 episodes are omitted from the visualization.
The number of evaluated episodes is shown for each category.}
    \label{fig:category_metrics}
\end{figure*}

\subsection{Additional Results and Analysis}
\noindent\textbf{Performance by Goal Category.}
Fig.~\ref{fig:category_metrics} shows that performance varies considerably by target category. On HM3D-v1 and v2, chairs, beds, and couches are found more reliably, while televisions and potted plants remain more difficult. Couches also have relatively high SPL, likely because they are large and tend to appear in predictable room types. We see a similar pattern on OVON: refrigerators and pianos are among the easier targets, while smaller or less visually distinctive objects such as hangers, blinds, and flowerpots are harder to locate. For clarity, we only show OVON categories with at least 25 episodes.

\begin{figure*}[h]
    \centering
    \includegraphics[width=\linewidth]{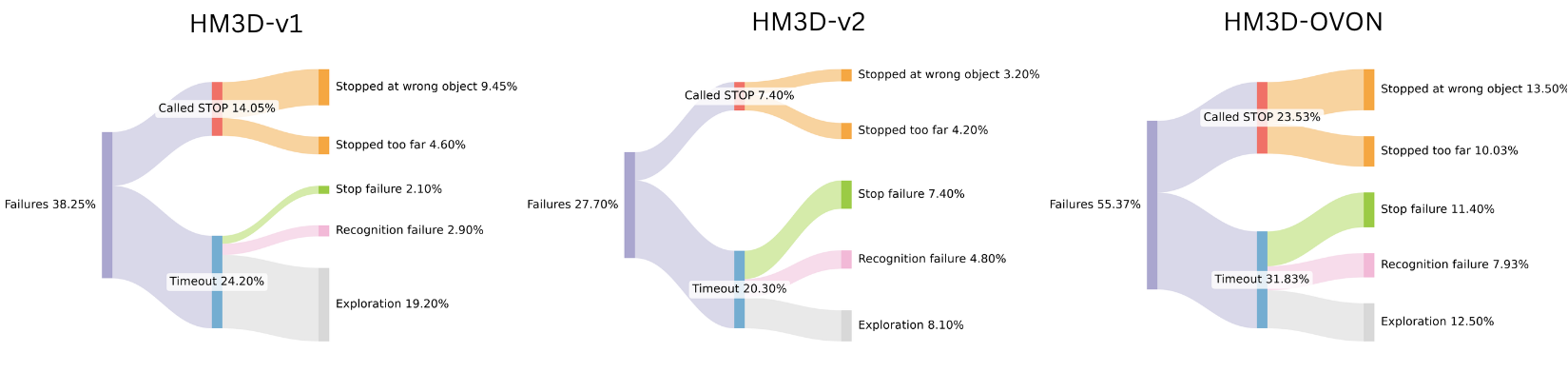}
    \caption{\textbf{RTNav failure modes} across HM3D-v1, HM3D-v2, and HM3D-OVON. We categorize failures into incorrect \texttt{STOP} decisions and timeouts, with their corresponding causes.}
    \label{fig:failure_analysis}
\end{figure*}

\noindent \textbf{Failure Analysis}
Fig.~\ref{fig:failure_analysis} breaks down RTNav's failure modes across the three benchmarks. On HM3D-v1 and HM3D-v2, most failures come from timeouts rather than incorrect \texttt{STOP} decisions. HM3D-v1 has more exploration failures, in part because its scenes can span multiple floors, so the target may lie on another floor or be missed during exploration. These cases are less common on the single-floor HM3D-v2 benchmark. Failures on HM3D-OVON are more spread out across categories, which is consistent with its more challenging open-vocabulary targets. The Stretch camera also has a narrower, taller field of view, making some objects easier to miss while exploring.

\subsection{Real-World Experiment Details}

\paragraph{Setup} We used the Hello Robot Stretch 3 for our real-world experiments. Object navigation tasks assume access to synchronized RGB-D, GPS, and yaw streamed from the Habitat simulation. In the real world, we streamed RGB-D directly from the head-mounted RealSense D435i camera, while the global pose was obtained by running SLAM onboard using the stock 2D-LiDAR mounted on the base. Specifically, we ran HectorSLAM from the HomeRobot repository. We synchronized these topics with 0.02 slop. All RTNav modules ran on a single NVIDIA Jetson Thor, which communicated with the robot over Wi-Fi. We evaluated RTNav on 16 target objects across two indoor areas of an academic building, including classrooms, a kitchen, restrooms, a common area, and connecting hallways.

\paragraph{Real-World Specific Changes}

We used the RTNav modules described in Section~\ref{sec:method_agent} with several changes for real-world deployment. For mapping, we combined the SLAM occupancy map with obstacles reconstructed from RGB-D. The SLAM map was restricted to the region within the camera field of view and depth range and fused with voxelized 3D points obtained from the depth observations. The depth projection captured obstacles above the height of the 2D LiDAR, while the SLAM map provided a more stable representation for long-horizon navigation. We additionally applied temporal and spatial filtering to reduce noise caused by depth and pose errors. The resulting 2D map was used to compute frontiers. We adjusted frontier hyperparameters to the scale of each environment.

In simulation, we used a PointNav policy to remain consistent with common baselines for the Habitat ObjectNav tasks. Because PointNav is trained on the Habitat-specific visual distribution, we did not deploy it directly in the real world. Instead, we replaced it with a modified version of the open-sourced FMM planner from L3MVN, which generated waypoints to the selected goal. A continuous PD controller then tracked these waypoints. The remaining reasoning, detection, and verification components were directly applied without modification. 

%% file: rtnav/tables/action_map.tex
\begin{table}[h]
\centering
\caption{Discrete-to-continuous action mapping: 1 sync step = 1s.}
\label{tab:action_mapping}
\small
\begin{tabular}{l c c}
\toprule
Action & Command & Termination \\
\midrule
\textsc{MoveForward}              & $v = 0.25\ \mathrm{m/s}$     & $\Delta d = 0.25\ \mathrm{m}$   \\
\textsc{RotateL / RotateR} & $\omega = \mp 30^{\circ}\!/\mathrm{s}$      & $\Delta\theta = \mp 30^{\circ}$ \\
\textsc{LookUp / LookDown}        & $\dot{\phi} = \pm 30^{\circ}\!/\mathrm{s}$ & $\Delta\phi = \pm 30^{\circ}$   \\
\addlinespace[2pt]
\textsc{Stop}                     & \multicolumn{2}{c}{episode-complete signal} \\
\textsc{Idle}                     & \multicolumn{2}{c}{$(v,\,\omega,\,\dot{\phi}) = (0,\,0,\,0)$} \\
\bottomrule
\end{tabular}
\end{table}

%% file: rtnav/tables/reprod.tex
\begin{table*}[h]
\caption{Comparison between reproduced and reported results. SR = Success Rate, SPL = Success weighted by Path Length, DTG = Distance to Goal.}
\vspace{\baselineskip}
\centering
\scalebox{1.05}{%
\small
\begin{tabular}{lllcccccc}
\toprule
Benchmark & Method & Venue & \multicolumn{2}{c}{SR $\uparrow$} & \multicolumn{2}{c}{SPL $\uparrow$} & \multicolumn{2}{c}{DTG $\downarrow$}  \\
\cmidrule(lr){4-5} \cmidrule(lr){6-7} \cmidrule(lr){8-9}
& & & report. & reprod. & report. & reprod. & report. & reprod.\\
\midrule
\multirow{1}{*}{OVON}
& VLFM~\cite{vlfm} & \venue{ICRA'24}
& $35.2$ & $39.8$
& $19.6$ & $24.2$
& $-$ & $3.50$ \\
\midrule
\multirow{6}{*}{HM3D-v1}
& L3MVN~\cite{l3mvn} & \venue{IROS'23}
& $50.4$ & $47.9$
& $23.1$ & $22.6$
& $4.43$ & $4.62$ \\
& VLFM~\cite{vlfm} & \venue{ICRA'24}
& $52.5$ & $52.6$
& $30.4$ & $30.4$
& $-$ & $4.12$ \\
& OpenFMNav~\cite{openfmnav} & \venue{NAACL-F'24}
& $54.9$ & $46.3$
& $24.4$ & $20.7$
& $-$ & $4.78$ \\
& Trihelper~\cite{trihelper} & \venue{IROS'24}
& $56.5$ & $52.7$
& $25.3$ & $23.7$
& $3.87$ & $4.19$ \\
& GAMap~\cite{gamap} & \venue{NeurIPS'24}
& $53.1$ & $48.8$
& $26.0$ & $23.6$
& $-$ & $4.58$ \\
& BeliefMapNav~\cite{beliefmapnav} & \venue{NeurIPS'25}
& $61.4$ & $65.5$
& $30.6$ & $32.2$
& $-$ & $2.95$ \\
\midrule
\multirow{6}{*}{HM3D-v2}
& L3MVN~\cite{l3mvn} & \venue{IROS'23}
& $36.3$ & $58.9$
& $15.7$ & $24.9$
& $-$ & $2.51$ \\
& VLFM~\cite{vlfm} & \venue{ICRA'24}
& $63.6$ & $64.0$
& $32.5$ & $32.7$
& $-$ & $2.01$ \\
& OpenFMNav~\cite{openfmnav} & \venue{NAACL-F'24}
& $-$ & $57.5$
& $-$ & $23.8$
& $-$ & $2.43$ \\
& Trihelper~\cite{trihelper} & \venue{IROS'24}
& $-$ & $64.0$
& $-$ & $26.0$
& $-$ & $1.98$ \\
& GAMap~\cite{gamap} & \venue{NeurIPS'24}
& $-$ & $61.3$
& $-$ & $27.1$
& $-$ & $2.32$ \\
& BeliefMapNav~\cite{beliefmapnav} & \venue{NeurIPS'25}
& $-$ & $73.5$
& $-$ & $33.6$
& $-$ & $1.79$ \\
\bottomrule
\end{tabular}%
}
\label{tab:reprod}
\end{table*}